\documentclass[10pt,twocolumn,letterpaper]{article}

\usepackage{iccv}
\usepackage{times}
\usepackage{epsfig}
\usepackage{graphicx}
\usepackage{amsmath}
\usepackage{amssymb}
\usepackage[accsupp]{axessibility}  

\usepackage[breaklinks=true,bookmarks=false]{hyperref}

\iccvfinalcopy 

\def\iccvPaperID{****} 

\usepackage[capitalize]{cleveref}
\crefname{section}{Sec.}{Secs.}
\Crefname{section}{Section}{Sections}
\Crefname{table}{Table}{Tables}
\crefname{table}{Tab.}{Tabs.}

\newcommand*\samethanks[1][\value{footnote}]{\footnotemark[#1]}

\ificcvfinal\fi

\begin{document}

\title{SVQNet: Sparse Voxel-Adjacent Query Network \\ for 4D Spatio-Temporal LiDAR Semantic Segmentation}

\author{
\centerline{
Xuechao Chen\textsuperscript{\rm 1}\thanks{Equal contribution} \quad
Shuangjie Xu\textsuperscript{\rm 2} \samethanks \quad
Xiaoyi Zou\textsuperscript{\rm 3} \quad
Tongyi Cao\textsuperscript{\rm 3} \quad
Dit-Yan Yeung\textsuperscript{\rm 2} \thanks{Corresponding authors} \quad
Lu Fang\textsuperscript{\rm 1} \samethanks \quad
} \\
\centerline{
\textsuperscript{\rm 1}Tsinghua University \quad
\textsuperscript{\rm 2}Hong Kong University of Science and Technology\quad
\textsuperscript{\rm 3}Deeproute.ai}
}

\maketitle
\ificcvfinal\thispagestyle{empty}\fi

\begin{abstract}
   LiDAR-based semantic perception tasks are critical yet challenging for autonomous driving. Due to the motion of objects and static/dynamic occlusion, temporal information plays an essential role in reinforcing perception by enhancing and completing single-frame knowledge. Previous approaches either directly stack historical frames to the current frame or build a 4D spatio-temporal neighborhood using KNN, which duplicates computation and hinders real-time performance. Based on our observation that stacking all the historical points would damage performance due to a large amount of redundant and misleading information, we propose the Sparse Voxel-Adjacent Query Network (SVQNet) for 4D LiDAR semantic segmentation. To take full advantage of the historical frames high-efficiently, we shunt the historical points into two groups with reference to the current points. One is the Voxel-Adjacent Neighborhood carrying local enhancing knowledge. The other is the Historical Context completing the global knowledge. Then we propose new modules to select and extract the instructive features from the two groups. Our SVQNet achieves state-of-the-art performance in LiDAR semantic segmentation of the SemanticKITTI benchmark and the nuScenes dataset.
\end{abstract}

\section{Introduction}
\label{sec:myintro}

Serving as a robust 3D perception solution, LiDAR-based perception is under enthusiastic exploration by researchers, among which 3D LiDAR semantic segmentation, aiming at assigning a category label to each point in the whole LiDAR scene at the semantic level, is of great significance in autonomous driving and robotics.
Recently, a large number of literature~\cite{10.1007/978-3-030-58604-1_41, ye2021drinet, zhang2018context, qiu2021semantic, Hou_2022_CVPR, cpgnet} concentrates on semantic segmentation within a single frame.
However, the information in a single frame is affected by multiple factors: 1) occlusion problems caused by obstacles or the movement of ego-car, leading to incomplete information of the occluded objects; and 2) ambiguity between similar point clusters, for example, the fence looks similar to one side of a big truck, which severely degrades the performance of single-frame based LiDAR semantic segmentation.

\begin{figure}[t]
  \vspace{-10px}
  \centering
  \includegraphics[width=1.0\linewidth]{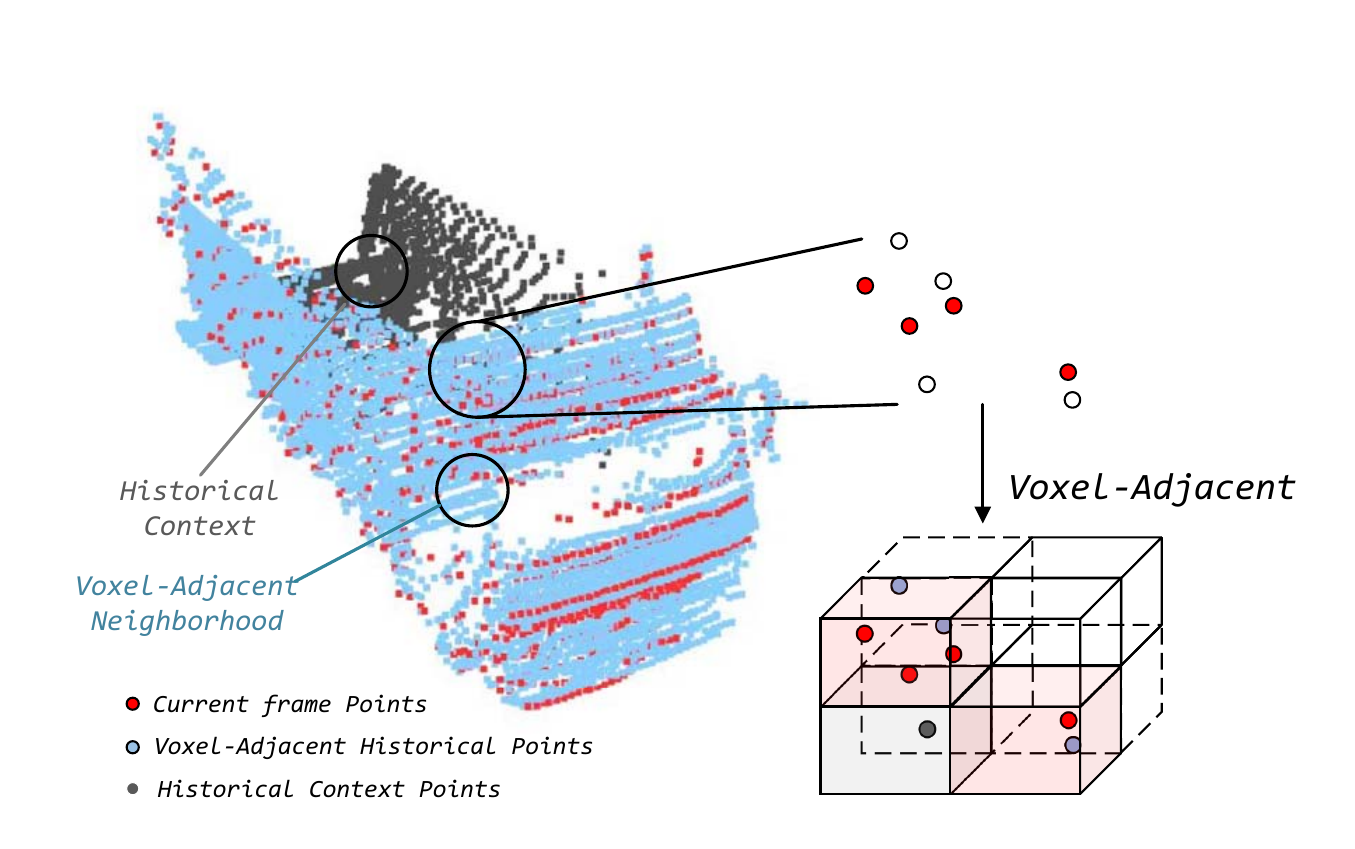}
  \caption{4D spatio-temporal LiDAR points for a truck. We shunt the historical points into 1) \textit{Voxel-Adjacent} points that lie in the voxel containing current frame points; 2) The remaining points named \textit{Historical Context} points whose features will be adaptively fused to complete the missing features in the current frame.}
   \label{fig:intro}
   
\end{figure}

  

To eliminate the distortion within the single frame, the use of sequential knowledge has attracted widespread attention~\cite{mccrae20203d, yuan2021temporal, ye2021tpcn, aygun20214d, wu2020motionnet} as the LiDAR continuously transmits and receives sensory data. Therefore, 4D Spatio-temporal information (LiDAR video) is increasingly playing an essential role in reinforcing perception by enhancing and completing single-frame knowledge. 
Classic temporal methods~\cite{aygun20214d, zhu2021cylindrical} directly stack frames in the last few timestamps by adding additional channel $t$ to the coordinates $xyz$ of each point, which is quite straightforward but superimposing all historical points without any selection brings redundancy, masking the useful temporal knowledge and weakens the benefits of the time series. 

To model spatio-temporal relationship instead of stacking all frames, approaches based on KNN or radius neighbors query ~\cite{Liu_2019_ICCV, cao2020asapnet, Shi_2020_CVPR, wang2021anchorbased} apply point-wise nearest neighbor search methods to extract instructive features across time and space. However, these approaches will not only fail when the target object is moving at high speed but also bear the high complexity of searching algorithms that lead to the inability to adopt long time-series information. Other approaches based on RNN~\cite{huang2020lstm, ercelik_temp-frustum_2021, kumar_real-time_2020} or memory~\cite{duerr2020lidar, lu2020monet} using a recurrent neural network or sequence-independent storage to memorize the instructive features from historical frames, can model the long sequence knowledge. Nevertheless, these approaches are unable to align recurrent features under sparse representation and thus adopt a range-image view~\cite{9197193, wang2022meta}, which are impossible to gain from the sparse representation~\cite{graham20183d} of point clouds. 

To efficiently extract valuable spatio-temporal features in 3D voxel representation, we propose a Sparse Voxel-Adjacent Query Network (\textit{SVQNet}). Our \textit{SVQNet} shunts the historical information into two groups based on the observation from Fig.~\ref{fig:intro}:
1) \textit{Voxel-Adjacent Neighborhood}: historical points around points in current frame can \textit{enhance} the spatial semantic features from sparse to dense across time to disambiguate current frame semantics; 2) \textit{Historical Context}: some occlusion 
can be \textit{completed} from multiple frames in a learning manner, by activating valuable historical context according to current voxel features.
Unlike previous work~\cite{Shi_2020_CVPR, Liu_2019_ICCV} requiring the calculation and sorting of distances between current and historical points to find nearest neighbors, the search of \textit{Voxel-Adjacent} is highly efficient, which performs sparse hash query from current points to historical points, under several scales from small to big, acquiring spatio-temporal neighbors from near to far.
The sparse hash algorithm allows us to reduce the complexity from quadratic to linear, which further endows us with real-time performance.
The proposed \textit{SVQNet} achieves state-of-the-art performance on SemanticKITTI \cite{Behley_2019_ICCV} and nuScenes \cite{Caesar_2020_CVPR} datasets while maintaining a real-time runtime.
Our main contributions are as followed:

\begin{itemize}
\setlength{\itemsep}{0pt}
\setlength{\parsep}{0pt}
\setlength{\parskip}{0pt}
    \item The Spatio-temporal information is formulated as \textit{enhancing} and \textit{completing} in the first time, with a novel \textit{Spatio-Temporal Information Shunt} module to efficiently shunt the stream of historical information.
    \item An efficient \textit{Sparse Voxel-Adjacent Query} module is proposed to search instructive neighbors in 4D sparse voxel space, and extract knowledge from the \textit{Voxel-Adjacent Neighborhood}.
    \item The learnable \textit{Context Activator} is introduced to activate and extract historical \textit{completing} information.
    \item We furthermore introduce a lightweight \textit{Temporal Feature Inheritance} method to collect features of historical frames and reuse them in the current frame.
\end{itemize}

\section{Related Work}
\label{sec:rela}


\subsection{LiDAR Semantic Segmentation}
\label{2.1}

3D semantic segmentation is to classify each LiDAR point with a semantic label.
Early work is mainly based on the indoor semantic dataset.
Pointnet \cite{Qi_2017_CVPR} treated point cloud data as one-dimensional and directly applied MLPs to extract features.
RandLA-Net \cite{Hu_2020_CVPR} proposed a local feature aggregation method based on K Nearest Neighbor (KNN).
KPConv \cite{Thomas_2019_ICCV} proposed a novel convolution on point clouds, which is called kernel point convolution.
Others \cite{Choy_2019_CVPR, graham20183d} employed sparse convolution which significantly speeds up 3D convolution and improves the performance. 

In recent years, with the advent of outdoor datasets~\cite{Behley_2019_ICCV,Caesar_2020_CVPR}, more and more 3D segmentation methods for large scenes are proposed.
JS3C-Net \cite{Yan_Gao_Li_Zhang_Li_Huang_Cui_2021} utilized shape priors from the scene completion task to help semantic segmentation.
SPVNAS \cite{10.1007/978-3-030-58604-1_41} proposed a two-branch Point-Voxel convolution method to extract sparse features of point clouds and a method of automatically searching the best model construct, which is called Neural Architecture Search.
DRINet \cite{ye2021drinet} proposed a dual representation that contains Point-Voxel and Voxel-Point feature extraction.
Some research transferred Cartesian coordinates to polar coordinates \cite{Zhang_2020_CVPR} and cylindrical coordinates \cite{zhu2021cylindrical}.
However, these methods only use features from a single frame and lack context information in temporal space.

\begin{figure*}
  \vspace{-10px}
    \centering
    \includegraphics[width=1.0\textwidth]{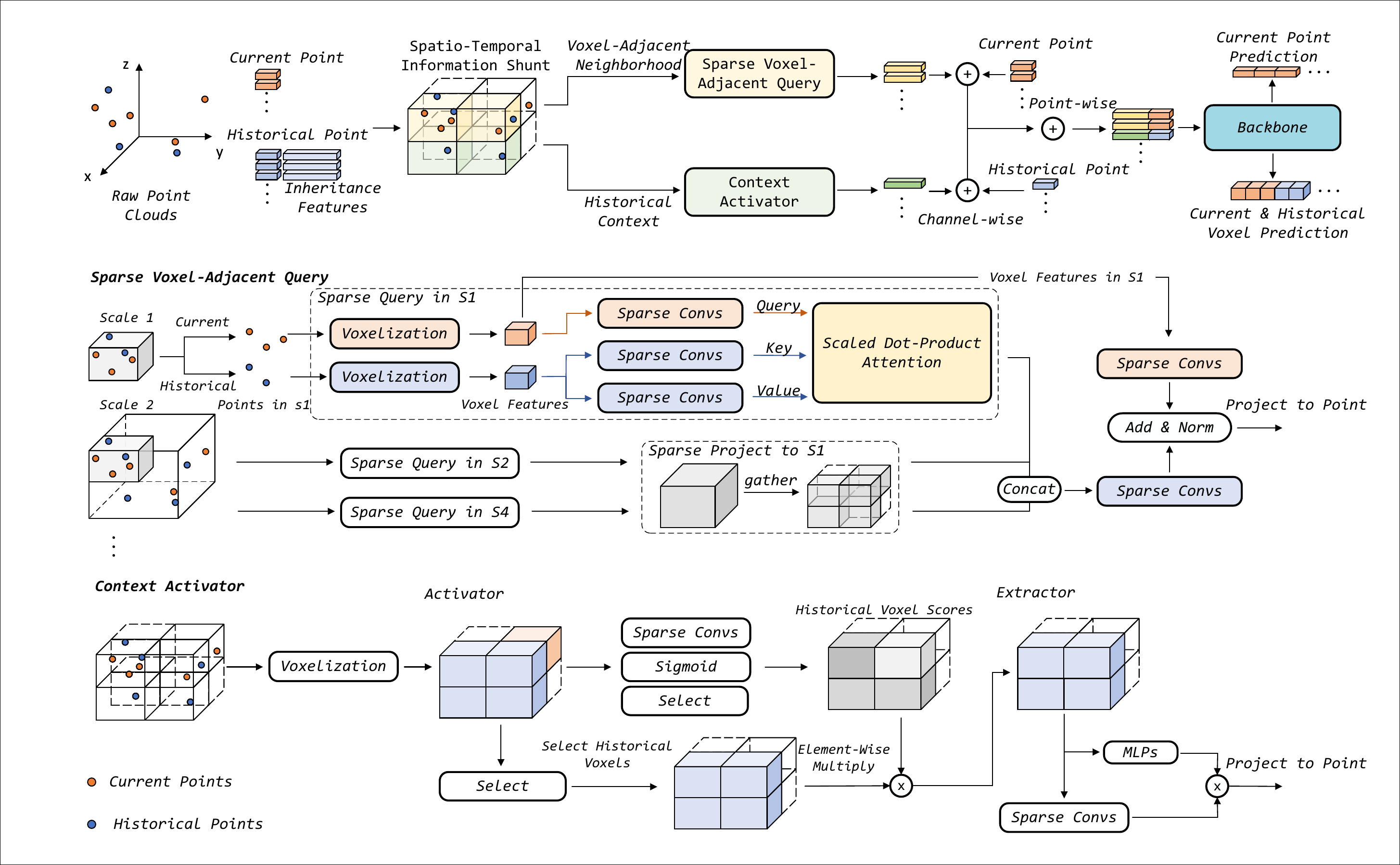}
    \caption{
    The architecture of our proposed \textit{SVQNet}.
     \textit{Spatio-Temporal Information Shunt} (\textit{STIS}) shunts historical sequences into \textit{Voxel-Adjacent Neighborhood} and \textit{Historical Context} information high-efficiently. 
     The shunted streams are then fed into \textit{Sparse Voxel-Adjacent Query} (\textit{SVAQ}) module and \textit{Context Activator} (\textit{CA}) module, where the former aggregates multi-scale voxel features from historical \textit{Voxel-Adjacent Neighborhood} to enhance current features with self-attention, and the latter activates \textit{Historical Context} features with a learnable scoring strategy to dynamic select the global context that is truly instructive.
     Finally, the output attentive features from the two modules are concatenated and fed into our backbone.
    }
    \label{fig3}
    \vspace{-5px}
  \end{figure*}

\subsection{Temporal LiDAR Perception}
\label{2.2}

Temporal information, namely 4D Spatio-temporal information, is usually considered useful information in LiDAR perception. 
Recently, lots of work made efforts on temporal LiDAR detection \cite{huang2020lstm, mccrae20203d, yuan2021temporal, kumar_real-time_2020}, motion prediction \cite{wu2020motionnet, ye2021tpcn}, scene flow estimation \cite{rishav2020deeplidarflow, puy2020flot, liu2019flownet3d}.
As for temporal LiDAR semantic segmentation, it aims to utilize 4D Spatio-temporal information to improve the performance of semantic segmentation.
SpSequenceNet \cite{Shi_2020_CVPR} employed KNN to gather 4D Spatio-temporal information globally and locally.
4D MinkNet \cite{Choy_2019_CVPR} directly processed 4D spatio-temporal point clouds using high-dimensional convolutions.
Some work \cite{cao2020asapnet, lu2020monet, schutt2022abstract} recurrently utilized sequential information.
DeepTemporalSeg \cite{9197193} employed dense blocks and utilized depth separable convolutions to explore the temporal consistency.
LMNet~\cite{chen2021moving} effectively performs scene segmentation by accurately distinguishing moving and static objects. 
4DMOS~\cite{mersch2022receding} utilizes sparse 4D convolutions to extract spatio-temporal features and predict moving object scores for each point.
Previous work can be summarized into three categories: directly frame stacking~\cite{aygun20214d, kreuzberg20234d}, KNN based~\cite{Shi_2020_CVPR,Choy_2019_CVPR}, RNN/memory based~\cite{cao2020asapnet, lu2020monet, schutt2022abstract}.
Nevertheless, the methods mentioned above not only lack the mining of the effective part of the 4D information but also suffer from the high computation of 4D processing.
In contrast, our method takes full advantage of the historical frames high-efficiently owing to the shunt of \textit{Voxel-Adjacent Neighborhood} and \textit{Historical Context}.

\section{Method}
\label{sec:method}


\noindent \textbf{Architecture.} 
As shown in \cref{fig:intro}, we shunt the historical points into two phases:
1) By modeling \textit{Voxel-Adjacent Neighborhood}, we enhance the current points using adjacent historical points.
2) The non-adjacent historical points that we call \textit{Historical Context} help complete the current scan. 
And we show the modeling processes of the two phases in \cref{3.1}.
Then the divided two streams are fed into \textit{Sparse Voxel-Adjacent Query} (detailed in \cref{3.2}) and \textit{Context Activator} (detailed in \cref{3.3}) respectively, as pipeline demonstrated in the top row of Fig.~\ref{fig3}.

\noindent \textbf{Plugin properties.} Shown as Fig.~\ref{fig3}, the proposed approach is a kind of ``Plugin" which can be used as a plugin to be inserted in mainstream backbone networks, making those networks able to take advantage of temporal information. The experiment on other methods is detailed in \cref{4.4}.

\subsection{Spatio-Temporal Information Shunt}
\label{3.1}

As shown in \cref{fig3}, the role of \textit{Spatio-Temporal Information Shunt} (\textit{STIS}) is to 
model the \textit{Voxel-Adjacent Neighborhood} and \textit{Historical Context} swiftly.
Specifically, \textit{STIS} takes in both current points $P^c=\{p_i^c,i=1,...,N\}$ where $p_i^c\in\mathbb{R}^5$ and historical points $P^h=\{p_j^h,j=1,...,M\}$ where $p_j^h\in\mathbb{R}^{5+d}$. 
The original point features contain $x, y, z$, intensity, and timestamp. 
$d$ represents the dimension of inheritance features from historical frames later detailed in \cref{3.4}.
All historical coordinates $x, y, z$ have been converted in the current coordinate system by translation and rotation according to the pose matrix of ego-motion.

\noindent \textbf{Voxelization} is to divide 3D space into voxels with size $w,l,h$ and then assign each point to the voxel it lies in. In practice, the point clouds will be projected to voxels at different scales $s$ (a positive number), which means that the voxel size will become $s \times w,s \times l,s \times h$. Therefore, the voxelized coordinate for a point under scale $s$ is $\left\lfloor {\frac{x}{{s \times w}}} \right\rfloor ,\left\lfloor {\frac{y}{{s \times l}}} \right\rfloor ,\left\lfloor {\frac{z}{{s \times h}}} \right\rfloor$. Given point clouds $P$, we can apply voxelization to get voxelized coordinates of all points, and these unique voxelized coordinates form the voxel set $V^s = \left\{ {{v^s_k},k = 1,...,L} \right\}$. Each ${v^s_k}$ contains a voxelized coordinate and the corresponding features for this voxel ${f^s_k}$, which is aggregated from point features inside the voxel by applying DynamicVFE~\cite{zhou2020end}.
To simplify symbols, we hide $s$ if all variables in one formula are in the same scale.

\noindent \textbf{Voxel-Adjacent Neighborhood Modeling.}
Previous attempts~\cite{cao2020asapnet, wang2021anchorbased, Shi_2020_CVPR, Liu_2019_ICCV} have proved that building a local spatio-temporal neighborhood can enhance the performance in sequential perception.
Unlike previous KNN or radius-based methods \cite{Shi_2020_CVPR, Liu_2019_ICCV}, we model our proposed \textit{Voxel-Adjacent Neighborhood} with real-time inference speed.
\textit{Voxel-Adjacent Neighborhood} search is to query historical voxel features whose voxelized coordinates also exist in the current frame.
The physical meaning is to use the method of coincidence of voxelized coordinates to quickly find the nearest neighbor of the current points from the historical points.
Given current point clouds $P^c$ and historical point clouds $P^h$ whose points have been projected into the coordinate system of $P^c$, we can get the corresponding current voxel set $V^c$ and historical voxel set $V^h$ by voxelization under scale $s$.
Then the queried historical voxel set $V^q$ can be obtained by the $\Psi _{{\rm{query}}}$ function:
\begin{equation}
\label{eq:sparsehashquery}
\begin{aligned}
{V^q} &= {\Psi _{{\rm{query}}}}({V^h},{V^c})\\
 &= {\rm{HashQuery}}\left( {{\cal C}\left( {{V^c}} \right),{\cal C}\left( {{V^h}} \right)} \right).
\end{aligned}
\end{equation}
$\mathcal{C}$ is the function to get the voxelized coordinate if input one voxel or a set of coordinates if input a set of voxels. Using the coordinates as hash keys, we can use the sparse $\mathop{\rm HashQuery}$ function to map $V^h$ to $V^c$:
\begin{equation}
\label{eq:corresponding_map}
    \mathop{\rm HashQuery}\left( {v_i^{c}}, {V^h} \right)=\left\{
    \begin{aligned}
    \emptyset & , & \mathcal{C}\left( {v_i^{c}} \right) \notin \mathcal{C}\left( {V^h} \right), \\
    v_j^{h} & , &  \mathcal{C}\left( {v_i^{c}} \right) = \mathcal{C}\left( {v_j^{h}} \right).
    \end{aligned}
    \right.
\end{equation}
$\emptyset$ denotes if there's no voxel under the corresponding coordinate in $V^h$, a zero placeholder will be padded in the query results. Otherwise, $\mathop{\rm HashQuery}$ returns \textit{Voxel-Adjacent Neighbor} ${v_j^{h}}$ which has the same coordinate as the input ${v_i^{c}}$.
The previous methods employ KNN or radius-based search to find 4D neighbors and the time complexity is $O(NM)$ ($M>N$). Ours benefits from the hash algorithm, whose time complexity is $O(N)$. \texttt{TorchSparse}~\cite{tang2022torchsparse} is used as the implementation of sparse hash query in this paper.

By the same procedures under multiple scales, we can get the \textit{Voxel-Adjacent Neighborhood} from small scale to big, from small scope to large.
Thus we model a multi-scale voxel-aware neighborhood search procedure under multiple scales to obtain the queried historical voxel sets $\{V^{qs}|s=s_1,s_2,s_4\}$, where the $s_i=i$ that is $1, 2, 4$ respectively in our setting.
Note that $V^c$ has the same length as $V^q$ and they are index-corresponding.
Then the built \textit{Voxel-Adjacent Neighborhood} carries local 4D Spatio-temporal information, which is extracted as enhancing features to the current frame by \textit{SVAQ}. More details about \textit{SVAQ} are described in \cref{3.2}.


\noindent \textbf{Historical Context Modeling.}
With the motion of the ego vehicle, part of the scene may not be visible caused of occlusion from other dynamic agents or static obstacles, leading to the target object being too sparse to be recognized.
In this case, points in historical frames can help to complete the lost information by aggregating \textit{Historical Context} from multiple frames.
\textit{Historical Context} denotes those historical voxel features that are not queried as \textit{Voxel-Adjacent Neighborhood}.
To activate valuable contexts selectively, we first select historical voxels that are not queried, which is $V^n$, and then provide these features to the \textit{Context Activator} module for further \textit{completing} features extraction.
Similar to the modeling of \textit{Voxel-Adjacent Neighborhood}, the unqueried voxels are named \textit{Historical Context}, which can be modeled by a negative function to $\Psi_{\mathop{\rm query}\nolimits}$:
\begin{equation}
\label{eq:psin}
    V^n=\Psi_{\mathop{\rm unquery}\nolimits}(V^h, V^c),
\end{equation}
where $V^h$ is the historical voxels and $V^{c}$ is the current voxels. $V^n$ is the remaining elements of historical voxel set $V^h$ that are not queried by $\Psi_{\mathop{\rm query}\nolimits}$. The scale of \textit{Historical Context} $s_1=1$ in our setting.
Subsequently, the formed \textit{Historical Context} carrying global 4D Spatio-temporal information is fed to \textit{Context Activator} module detailed in \cref{3.3}.

\subsection{
Sparse Voxel-Adjacent Query}
\label{3.2}

The key to enhancing the current features is to extract the relative historical features with adaptive learning. Therefore, we propose a \textit{Sparse Voxel-Adjacent Query (SVAQ)} module to benefit from \textit{Voxel-Adjacent Neighborhood} with Transformer attention.
As illustrated in \cref{fig3}, given voxel set $\{V^c, V^q\}$ under scale $s$, voxel features from current frame $V^c$ are fed into one Sparse-Convolution \cite{Choy_2019_CVPR, graham20183d} layer (${\mathop{\rm SPC}\nolimits}$) to generate \textit{Query} features.
The queried historical voxels $V^q$ are fed into two independent Sparse-Convolution layers to generate \textit{Key} and \textit{Value} features.
Later the generated \textit{Query}, \textit{Key} of dimension $d_k$ and \textit{Value} are fed into Scaled Dot-Product Attention \cite{10.5555/3295222.3295349} to extract attentive local 4D spatio-temporal voxel features $T$ under scale $s$, which can be written as:
\begin{equation}
    T = {\mathop{\rm SPC}\nolimits}({V^q}) \cdot {\mathop{\rm Softmax}\nolimits}(\frac{{\mathop{\rm SPC}\nolimits}({V^c}) \cdot {\mathop{\rm SPC}\nolimits}({V^q})^T}{ \sqrt{d_k}}).
\end{equation}
$T$ contains local dependencies between current features $V^c$ and historical \textit{Voxel-Adjacent} features $V^q$, making the repetitive historical features more instructive to \textit{enhance} the same voxel in current frame.
Then the attentive features under three scales $\left\{ {{s_1},{s_2},{s_4}} \right\}$ extracted by Scaled Dot-Product Attention are concatenated at the feature channel and fused by another three ${\mathop{\rm SPC}\nolimits}$ layers.
Specially, the attentive features $T^{s_2}$, $T^{s_4}$ 
should be projected back to scale 1.
The calculation can be denoted as
\begin{equation}
    T_o = {\mathop{\rm SPC}\nolimits}( T^{s_1} \oplus {\mathop{\rm Proj}\nolimits}(T^{s_2}) \oplus {\mathop{\rm Proj}\nolimits}(T^{s_4}) ),
\end{equation}
where $\oplus$ denotes channel-wise concatenation, ${\mathop{\rm Proj}\nolimits}$ denotes sparse projection to scale $s_1=1$, which can be implemented as a $\mathop{\rm HashQuery}$ task by querying ${T^s}$ for $s>1$ according to the scale-normlized coordinates of ${T^{s_1}}$ to scatter features from low resolution to high resolution:
\begin{equation}
\label{eq:projection}
    {\mathop{\rm Proj}\nolimits}(T^{s}) = {\mathop{\rm HashQuery}\nolimits}\left( {{{\mathcal{C}\left( {{T^{{s_1}}}} \right)} \mathord{\left/
 {\vphantom {{\mathcal{C}\left( {{T^{{s_1}}}} \right)} s}} \right.
 \kern-\nulldelimiterspace} s},\mathcal{C}\left( {{T^s}} \right)} \right).
\end{equation}

In particular, we take $V^c$ under scale ${s_1}$ as a skip connection followed by three ${\mathop{\rm SPC}\nolimits}$ layers.
The final output features $O_v$ of \textit{SVAQ} can be represented as:
\begin{equation}
    O_v = {\mathop{\rm Prop}\nolimits}( {\mathop{\rm Norm}\nolimits}( {\mathop{\rm SPC}\nolimits}(V^{cs_1}) + T_o) ),
\end{equation}
where ${O_v} \in \mathbb{R}{^{N \times N_C}}$, $N$ is current point number and $N_C$ is the feature channel number. ${\mathop{\rm Prop}\nolimits}$ denotes the projection from voxels to points, ${\mathop{\rm Norm}\nolimits}$ denotes BatchNorm layer~\cite{ioffe2015batch}.
${\mathop{\rm Prop}\nolimits}$ can be formulated as the inversion of \textit{Voxelization}, which outputs point features by assigning the feature of each point with the voxel feature it lies in.
The number of the projected point features equals the number of the input current points and they are index-corresponding.
In summary, our \textit{SVAQ} module acts in a multi-head attention way to encode \textit{Voxel-Adjacent} features to current point features.

\begin{table*}[!t]
    \centering
    \resizebox{\textwidth}{!}{
\setlength{\tabcolsep}{1.3mm}{      \def\arraystretch{1.1}
    \begin{tabular}{@{}l | c c c c c c c c c c c c c c c c c c c c c c c c c | c @{}}
    \hline
    Methods & \rotatebox{90}{road} & \rotatebox{90}{sidewalk} & \rotatebox{90}{parking} & \rotatebox{90}{other ground} & \rotatebox{90}{building} & \rotatebox{90}{car} & \rotatebox{90}{truck} & \rotatebox{90}{bicycle} & \rotatebox{90}{motorcycle} & \rotatebox{90}{other vehicle \,} & \rotatebox{90}{vegetation} & \rotatebox{90}{trunk} & \rotatebox{90}{terrain} & \rotatebox{90}{person} & \rotatebox{90}{bicyclist} & \rotatebox{90}{motorcyclist} & \rotatebox{90}{fence} & \rotatebox{90}{pole} & \rotatebox{90}{traffic sign} & \rotatebox{90}{mov. car} & \rotatebox{90}{mov. bicyclist} & \rotatebox{90}{mov. person} & \rotatebox{90}{mov. motorcyc.} & \rotatebox{90}{mov. truck} & \rotatebox{90}{mov. other veh.} & \rotatebox{90}{mIoU ($\%$)}
    \\
    \hline \hline
    TangentConv\cite{tatarchenko2018tangent} & 83.9 & 64.0 & 38.3 & 15.3 & 85.8 & 84.9 & 21.1 & 2.0 & 18.2 & 18.5 & 79.5 & 43.2 & 56.7 & 1.6 & 0.0 & 0.0 & 49.1 & 36.4 & 31.2 & 40.3 & 1.1 & 6.4 & 1.9 & \textbf{42.2} & \textbf{30.1} & 34.1
    \\
    \hline
    DarkNet53Seg\cite{Behley_2019_ICCV} & 91.6 & 75.3 & 64.9 & 27.5 & 85.2 & 84.1 & 20.0 & 30.4 & 32.9 & 20.7 & 78.4 & 50.7 & 64.8 & 7.5 & 0.0 & 0.0 & 56.5 & 38.1 & 53.3 & 61.5 & 14.1 & 15.2 & 0.2 & 37.8 & 28.9 & 41.6
    \\
    \hline
    SpSequenceNet\cite{Shi_2020_CVPR} & 90.1 & 73.9 & 57.6 & 27.1 & 91.2 & 88.5 & 29.2 & 24.0 & 26.2 & 22.7 & 84.0 & 66.0 & 65.7 & 6.3 & 0.0 & 0.0 & 66.8 & 50.8 & 48.7 & 53.2 & 41.2 & 26.2 & 36.2 & 0.1 & 2.3 & 43.1
    \\
    \hline
    TemporalLidarSeg\cite{duerr2020lidar} & 91.8 & 75.8 & 59.6 & 23.2 & 89.8 & 92.1 & 39.2 & 47.7 & 40.9 & 35.0 & 82.3 & 62.5 & 64.7 & 14.4 & 0.0 & 0.0 & 63.8 & 52.6 & 60.4 & 68.2 & 42.8 & 40.4 & 12.9 & 2.1 & 12.4 & 47.0
    \\
    \hline
    KPConv\cite{Thomas_2019_ICCV} & 86.5 & 70.5 & 58.4 & 26.7 & 90.8 & 93.7 & \textbf{42.5} & 44.9 & 47.2 & 38.6 & 84.6 & 70.3 & 66.0 & 21.6 & 0.0 & 0.0 & 64.5 & 57.0 & 53.9 & 68.1 & 67.4 & 67.5 & 47.2 & 0.5 & 0.5 & 51.2
    \\
    \hline
    Cylinder3D\cite{zhu2021cylindrical} & 90.4 & 74.9 & 66.3 & 32.1 & 92.4 & 93.8 & 41.2 & \textbf{67.6} & \textbf{63.3} & 37.6 & 85.4 & 72.8 & 68.1 & 12.9 & \textbf{0.1} & \textbf{0.1} & 65.8 & 62.6 & 61.3 & 68.1 & 60.0 & 63.1 & 0.4 & 0.0 & 0.1 & 51.5
    \\
    \hline \hline
    \textbf{SVQNet}\textbf{(ours)} & \textbf{93.2} & \textbf{80.5} & \textbf{71.6} & \textbf{37.0} & \textbf{93.7} & \textbf{96.1} & 40.4 & 64.4 & 60.3 & \textbf{60.9} & \textbf{87.3} & \textbf{76.7} & \textbf{72.3} & \textbf{27.4} & 0.0 & 0.0 & \textbf{72.6} & \textbf{68.4} & \textbf{71.0} & \textbf{80.5} & \textbf{72.4} & \textbf{84.7} & \textbf{91.0} & 3.9 & 7.5 & \textbf{60.5}
    \\
    \hline
    Improvements $\Delta$ & \textbf{+1.4} & \textbf{+4.7} & \textbf{+5.3} & \textbf{+4.9} & \textbf{+1.3} & \textbf{+2.3} & -2.1 & -3.2 & -3.0 & \textbf{+22.3} & \textbf{+1.9} & \textbf{+3.9} & \textbf{+4.2} & \textbf{+5.8} & -0.1 & -0.1 & \textbf{+5.8} & \textbf{+5.8} & \textbf{+9.7} & \textbf{+12.3} & \textbf{+5.0} & \textbf{+17.2} & \textbf{+43.8} & -38.3 & -22.6 & \textbf{+9.0}
    \\
    \hline
    \end{tabular}}}
    \caption{The experiment results on the semantic segmentation of SemanticKITTI test set (multi-scan phase).
    All listed methods utilized 4D spatio-temporal information according to their paper.
    $\Delta$: comparing with the best previous results for each class.
    (mov. denotes moving.)}
    \label{25classes}
\end{table*}

\begin{table*}[!t]
    \small
    \centering
    \resizebox{0.75\textwidth}{!}{
\setlength{\tabcolsep}{1.3mm}{      \def\arraystretch{1.1}
    \begin{tabular}{@{}l c c c c c c c c c c c c c c c c | c c@{}}
    \hline
    Methods & \rotatebox{90}{barrier} & \rotatebox{90}{bicycle} & \rotatebox{90}{bus} & \rotatebox{90}{car} & \rotatebox{90}{con. vehicle} & \rotatebox{90}{motorcycle} & \rotatebox{90}{pedestrian} & \rotatebox{90}{traffic cone} & \rotatebox{90}{trailer} & \rotatebox{90}{truck} & \rotatebox{90}{surface} & \rotatebox{90}{other flat} & \rotatebox{90}{sidewalk} & \rotatebox{90}{terrain} & \rotatebox{90}{manmade} & \rotatebox{90}{vegetation} & \rotatebox{90}{mIoU ($\%$)}
    \\
    \hline \hline
    PolarNet\cite{Zhang_2020_CVPR} & 72.2 & 16.8 & 77.0 & 86.5 & 51.1 & 69.7 & 64.8 & 54.1 & 69.7 & 63.5 & 96.6 & 67.1 & 77.7 & 72.1 & 87.1 & 84.5 & 69.4
    \\
    \hline
    JS3C-Net\cite{Yan_Gao_Li_Zhang_Li_Huang_Cui_2021} & 80.1 & 26.2 & 87.8 & 84.5 & 55.2 & 72.6 & 71.3 & 66.3 & 76.8 & 71.2 & 96.8 & 64.5 & 76.9 & 74.1 & 87.5 & 86.1 & 73.6
    \\
    \hline
    Cylinder3D\cite{zhu2021cylindrical} & 82.8 & 29.8 & 84.3 & 89.4 & 63.0 & 79.3 & 77.2 & 73.4 & \textbf{84.6} & 69.1 & \textbf{97.7} & 70.2 & 80.3 & 75.5 & 90.4 & 87.6 & 77.2
    \\
    \hline
    SPVNAS\cite{10.1007/978-3-030-58604-1_41} & 80.0 & 30.0 & 91.9 & 90.8 & 64.7 & 79.0 & 75.6 & 70.9 & 81.0 & 74.6 & 97.4 & 69.2 & 80.0 & 76.1 & 89.3 & 87.1 & 77.4 
    \\
    \hline
    AF2S3Net\cite{cheng20212} & 78.9 & \textbf{52.2} & 89.9 & 84.2 & \textbf{77.4} & 74.3 & 77.3 & 72.0 & 83.9 & 73.8 & 97.1 & 66.5 & 77.5 & 74.0 & 87.7 & 86.8 & 78.3
    \\
    \hline \hline
    \textbf{SVQNet}\textbf{(ours)} & \textbf{84.5} & 41.8 & \textbf{93.3} & \textbf{92.5} & 69.1 & \textbf{85.5} & \textbf{83.7} & \textbf{78.3} & 84.5 & \textbf{77.5} & 97.1 & \textbf{70.3} & \textbf{81.6} & \textbf{77.9} & \textbf{91.8} & \textbf{90.1} & \textbf{81.2}
    \\
    \hline
    \end{tabular}}}
    \caption{Results on the nuScenes test set.
    All listed methods employed frames stacking strategy according to their implementation.}
    \label{nuScenes}
\end{table*}

\subsection{Context Activator}
\label{3.3}

Since the information in the current frame is incomplete caused of dynamic or static occlusion and the sparsity of LiDAR beams, we utilize 4D Spatio-temporal information to \textit{complete} current features with the automatically selected valuable \textit{Historical Context}.
Yet directly stacking historical points damages the efficiency and brings a lot of repetitive information, we propose learning-based \textit{Context Activator} (\textit{CA}) shown in \cref{fig3} to flexibly activate and extract the global \textit{Context} that are truly instructive. 
Firstly, the \textit{Activator} generates voxel scores $S$ for each context voxel of $V^n$, with the help of current voxels $V^c$ as reference.
It employs a three-layer ${\mathop{\rm SPC}\nolimits}$ along with a Sigmoid layer ${\mathop{\rm Sigmoid}\nolimits}$ to scoring $V^n$ with a predicted score between $0$ to $1$.
Then we multiply the \textit{Context} $V^n$ with corresponding voxel scores to perform an element-wise attentive selection with a defined threshold $S_{th}$.
Note that with $S_{th}$, the number of historical voxels reserved after the selection process can be controlled to balance the performance and computation at inference time.
The \textit{Activator} can be represented as:
\begin{equation}
    \begin{array}{l}
        S = {\mathop{\rm Select}\nolimits}_{V^n} \left( {\mathop{\rm Sigmoid}\nolimits}\left( {{\mathop{\rm SPC}\nolimits}\left( V^{n} \circ V^{c} \right)} \right) \right),
    \end{array}
\end{equation}
\begin{equation}
    \begin{array}{l}
        R = {\mathop{\rm Select}\nolimits}_{ S > {S_{th}}} \left(  V^{n} \otimes S \right),
    \end{array}
\end{equation}
where 
$\otimes$ denotes element-wise multiplication, $\circ$ denotes the union set of two voxel sets. The ${\mathop{\rm Select}\nolimits}_{V^{n}}$ selects scores generated by ${V^{n}}$.
The ${\mathop{\rm Select}\nolimits}_{ S > {S_{th}}}$ selects scores bigger than ${S_{th}}$, which is only activated at inference time. In the training process, the $R$ is obtained by $R = V^{n} \otimes S$ to keep samples in the low score as negative samples to balance the learning of \textit{Activator}.

After the dynamic selection, the \textit{Extractor} performs channel-wise self-attention on activated \textit{Historical Context}.
\textit{Extractor} employs a three-layer ${\mathop{\rm MLP}\nolimits}$ and a three-layer ${\mathop{\rm SPC}\nolimits}$ to extract features from $R$ respectively, where 
the former takes voxels as points to extract inner-voxel features and the latter captures the inter-voxel relationship.
Then the final output $O_c$ is obtained by the product of two groups:
\begin{equation}
    O_c = {\mathop{\rm Prop}\nolimits}( {\mathop{\rm MLP}\nolimits}(R) \cdot {\mathop{\rm SPC}\nolimits}(R) ),
\end{equation}
where ${O_c} \in \mathbb{R}{^{M' \times N_C}}$, $M'$ is the number of points in activated \textit{Historical Context} voxels and $N_C$ denotes channel number same as $O_v$.
To sum up, our \textit{CA} acts in a learnable way to activate \textit{Historical Context} selectively.

\begin{figure}[t]
  \centering
   \includegraphics[width=0.9\linewidth]{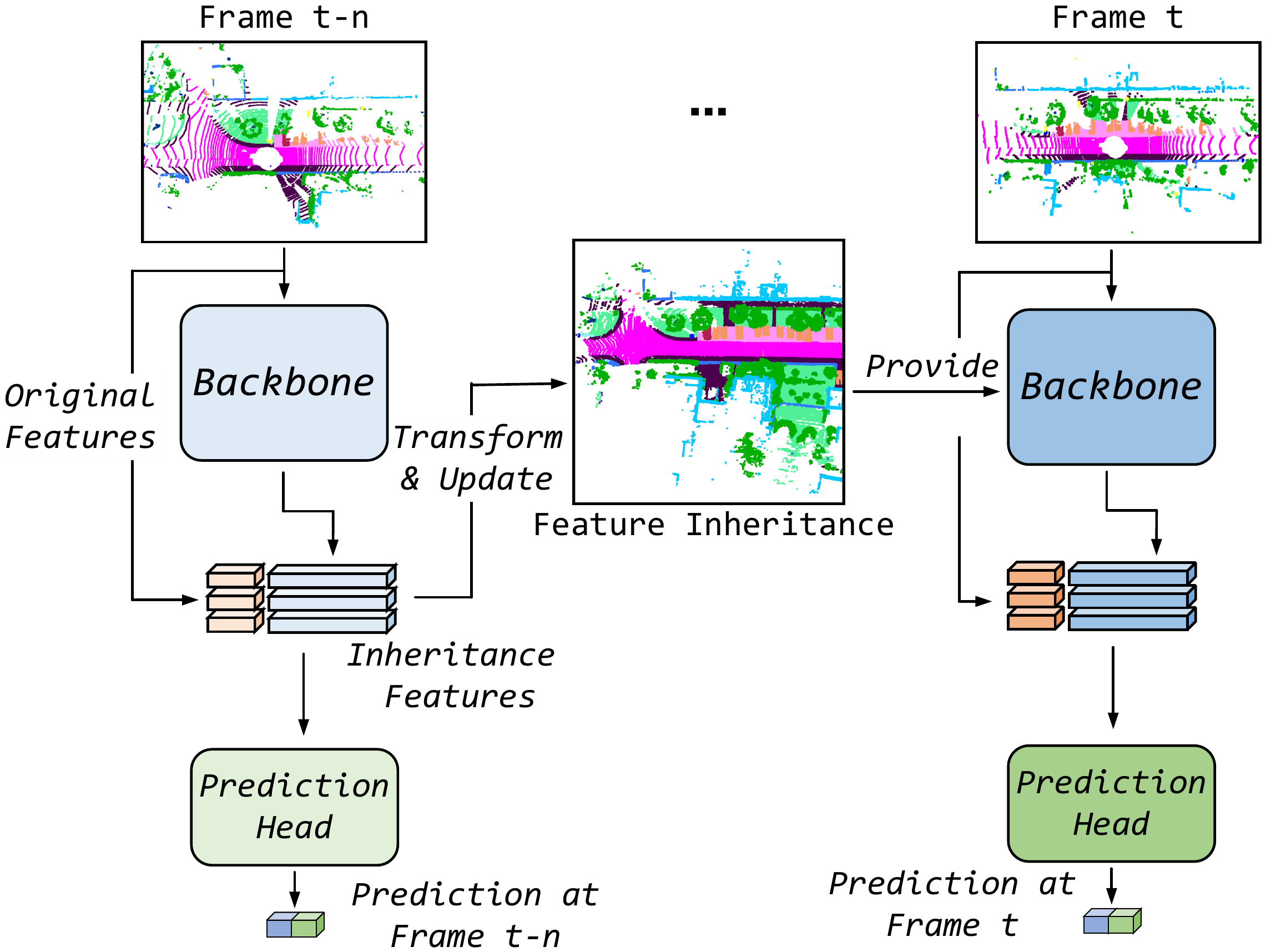} \caption{
   \textit{Temporal Feature Inheritance}. 
   We store the historical point features, namely inheritance features, along with the meta-information $x,y,z$, intensity, and timestamp. 
   Then we reuse them by transforming and updating coordinates in meta-information into coordinates under the current frame with the pose matrix.
   }
   \label{fig4}
\end{figure}

\subsection{Temporal Feature Inheritance}
\label{3.4}

We find that it is a waste to discard the high-dimensional features of previous frames and relearn them during the process of the current frame.
Therefore, as demonstrated in \cref{fig4}, we propose a simple but efficient method named \textit{Temporal Feature Inheritance} (\textit{TFI}) to inherit sparse features from historical sequences.
Assume that \textit{SVQNet} takes $n$ historical frames as input, all historical point features for frame $t$ (points from frame $t-n$ to $t-1$) are ready when the process of frame $t-1$ is finished by \textit{SVQNet}, which is like a sliding window on time series.
Based on the above observation, we concatenate high-dimensional point features from historical frames at the end of \textit{SVQNet} backbone with the corresponding meta-information including $x,y,z$, intensity, and timestamp, which is then stored in the buffer memory.
The historical meta-information is mainly used for the transformation of the coordinates of the points from the coordinate system of the historical frame into the current frame.
When processing the next frame $t$, we fetch historical point features from the memory and project $x,y,z$ to the coordinate system of the current frame by pose transformation, and then the projected coordinates can be used for \textit{SVAQ}.
Furthermore, the stored absolute timestamp should be transformed into a relative timestamp by subtraction with the absolute timestamp of the current frame.
At the end of \textit{SVQNet} at frame $t$, we update the memory with newly extracted features for the $t+1$ frame.


\section{Experiments}
\label{sec:exp}

\subsection{Datasets and Metrics}
\label{4.1}

\noindent \textbf{SemanticKITTI} \cite{Behley_2019_ICCV} is a large-scale outdoor point clouds dataset for autonomous driving which was collected by a $64$-beam LiDAR sensor.
The train set contains $23201$ sequential LiDAR scans and the test set contains $20351$ sequential scans.
The semantic segmentation task is officially divided into two phases.
One is the single-scan phase containing $19$ semantic classes without distinction of moving or static objects.
The other is the multi-scan phase containing $25$ semantic classes with the distinction between moving and static objects.
To test and ablate our \textit{SVQNet}, we run experiments on the multi-scan phase.

\noindent \textbf{nuScenes} \cite{Caesar_2020_CVPR, DBLP:journals/corr/abs-2109-03805} is a large-scale outdoor multi-modal dataset for autonomous driving.
Their point clouds data was collected by a $32$-beam LiDAR sensor in sequence.
It contains 1,000 scenes and 16 semantic classes with no distinction of moving or static objects.
They only annotated LiDAR data one frame every ten frames so the default setting of most methods is multiple scans.

\noindent \textbf{mIoU} is the mean intersection over the union.
The IoU is defined as $\frac{TP}{TP+FP+FN}$, where the
$TP$, $FP$, $FN$ represent the true positive, false positive, and false negative of the prediction.
The IoU score is first calculated for each class and then the mIoU is obtained by averaging across classes.

\noindent \textbf{Implementation details.}
We implement DRINet~\cite{ye2021drinet} as our baseline, and plus an additional voxel-wise loss to predict a semantic label for each voxel with the majority category of points in the voxel as the ground truth, which is an auxiliary loss to make \textit{CA} module to learn the activation for valuable historical voxels. Other settings are the same as DRINet reported. The point-wise loss is only applied to current points, but the voxel-wise loss is applied both to current voxels and historical voxels.

We adopt $1$ current frame and $2$ historical frames as the inputs without any down-sampling.
All the experiments are conducted on a machine with 8 * NVIDIA RTX 3090 GPU. The learning rate is set to $2e-3$ with an adamW optimizer~\cite{loshchilov2017decoupled}. The training epoch is set to $40$. 
The dimension $d$ of inheritance features, $d_k$ in \textit{SVAQ}, and $N_C$ in \textit{CA} are all set to 64.
The $S_{th}$ in \textit{CA} is set as $S_{th}=0.1$ to reserve activated historical voxel number $M' \approx 40k$ at inference time. In training, we set the $S_{th}=0.0$ to disable the selection.

\begin{figure}[t]
  \centering
  \includegraphics[width=1.0\linewidth]{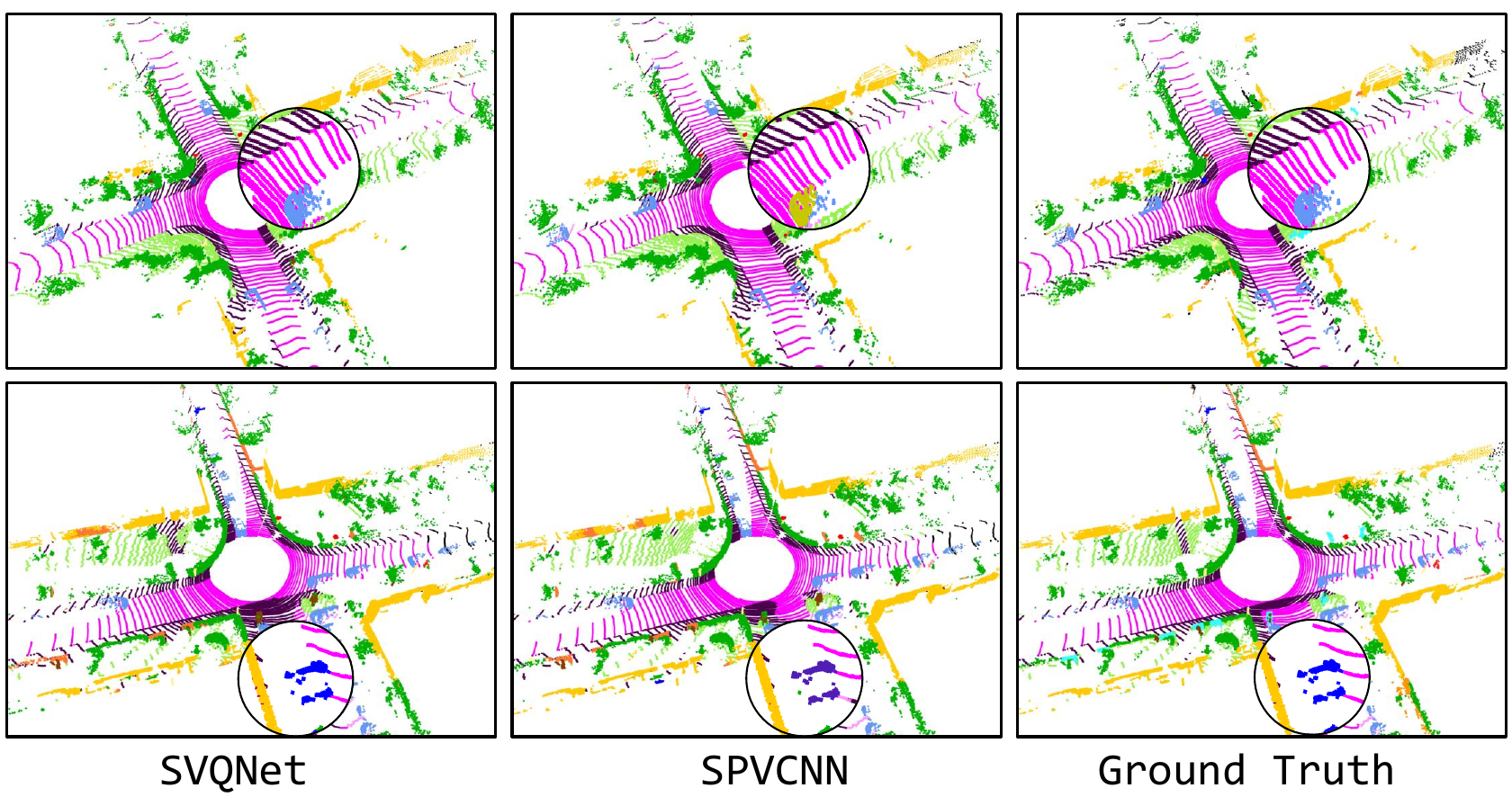} \caption{
  The qualitative visualization results on SemanticKITTI validation set, where the first column is the prediction from \textit{SVQNet}, the second column is the prediction from multi-scan SPVCNN~\cite{10.1007/978-3-030-58604-1_41} and the third column is the ground truth.
  Enlarged circles display our good cases (cars and motorcycles).
  }
  \label{fig5}
\end{figure}

\begin{table}
    \small
    \centering
    \resizebox{0.2\textwidth}{!}{
    \begin{tabular}{c|c}
    \hline
    Methods & Latency\\
    \hline \hline
    RandLA-Net~\cite{Hu_2020_CVPR} & 880 ms \\
    \hline
    SqueezeSegV3~\cite{xu2020squeezesegv3} & 238 ms \\
    \hline
    SPVNAS~\cite{10.1007/978-3-030-58604-1_41} & 259 ms \\
    \hline
    Cylinder3D~\cite{Zhu_2021_CVPR} & 170 ms \\
    \hline
    \textbf{SVQNet(ours)} & \textbf{97 ms} \\
    \hline
    \end{tabular}}
    \caption{
    Latency comparison with single-scan methods.
    }
    \vspace{-10 pt}
    \label{tab:runtime}
\end{table}

\begin{table}
\small
  \centering
  \scalebox{0.9}{
  \begin{tabular}{cccc|c|c|c}
    \hline
    Backbone+ & \textit{CA} & \textit{SVAQ} & \textit{TFI} & mIoU ($\%$) & $\Delta$ & Latency
    \\
    \hline \hline
    \checkmark &   &   &   & 52.8 & - & 125 ms \\
    \checkmark & \checkmark &   &   & 53.9 & \textbf{+1.1} & 61 ms \\
    \checkmark & \checkmark & \checkmark &   & 54.8 & +0.9 & 92 ms \\
    \checkmark & \checkmark & \checkmark & \checkmark & \textbf{55.3} & +0.5 & 97 ms \\
    \hline
  \end{tabular}}
  \caption{
  Ablation study on our proposed modules \textit{CA}, \textit{SVAQ}, and \textit{TFI}.
  Backbone+ denotes that the backbone network directly stacks $2$ historical frames with the current frame. Note that the Backbone+ takes in the same quantity of information as our \textit{SVQNet}.
  }
  \label{tab:ablation}
\end{table}

\begin{table}
\small
  \centering
  \begin{tabular}{c|ccccc}
    \hline
    $M'$ & 50$k$ & 45$k$ & 40$k$ & 35$k$ & 30$k$ \\
    \hline \hline
    mIoU ($\%$) & 55.0 & 55.2 & \textbf{55.3} & 55.1 & 54.9 \\
    \hline
  \end{tabular}
  \caption{
  Ablation study on \textit{Context Activator}.
  As the number of activated contextual voxels $M'$ varies by threshold $S_{th}$, \textit{SVQNet} achieves the best performance when $M'=40k$.
  }
  \label{tab:ablation3}
\end{table}

\subsection{Results on SemanticKITTI}
\label{4.2}

The multi-scan phase of SemanticKITTI distinguishes moving and static objects.
In this phase, we achieve state-of-the-art performance in terms of mIoU.
As shown in Tab.~\ref{25classes}, our proposed \textit{SVQNet} surpasses the multi-scan Cylinder3D \cite{zhu2021cylindrical} and KPConv \cite{Thomas_2019_ICCV}, which employed directly stacking strategy according to their implementation, by $9\%$ and $9.3\%$ in terms of mIoU separately.
We also obtain a $17.4\%$ performance gain in terms of mIoU than SpSequenceNet \cite{Shi_2020_CVPR}, which employed KNN to build the spatio-temporal relationship.
Besides, our SVQNet gets a mIoU improvement of $13.5\%$ compared with TemporalLidarSeg \cite{duerr2020lidar}, which proposed Temporal Memory Alignment to utilize spatio-temporal information.
The results show the excellent capability of our proposed \textit{SVQNet} in distinguishing moving and static semantics.
Due to the lack of reported runtime data of multi-scan methods listed in Tab.~\ref{25classes}, we compare the latency of our method with some 3D single-scan methods in Tab.~\ref{tab:runtime}, showing that our multi-scan method is even faster than previous single-scan methods.
Moreover, the qualitative visualization results are shown in Fig.~\ref{fig5}.

\begin{figure}[t]
  \centering
   \includegraphics[width=0.9\linewidth]{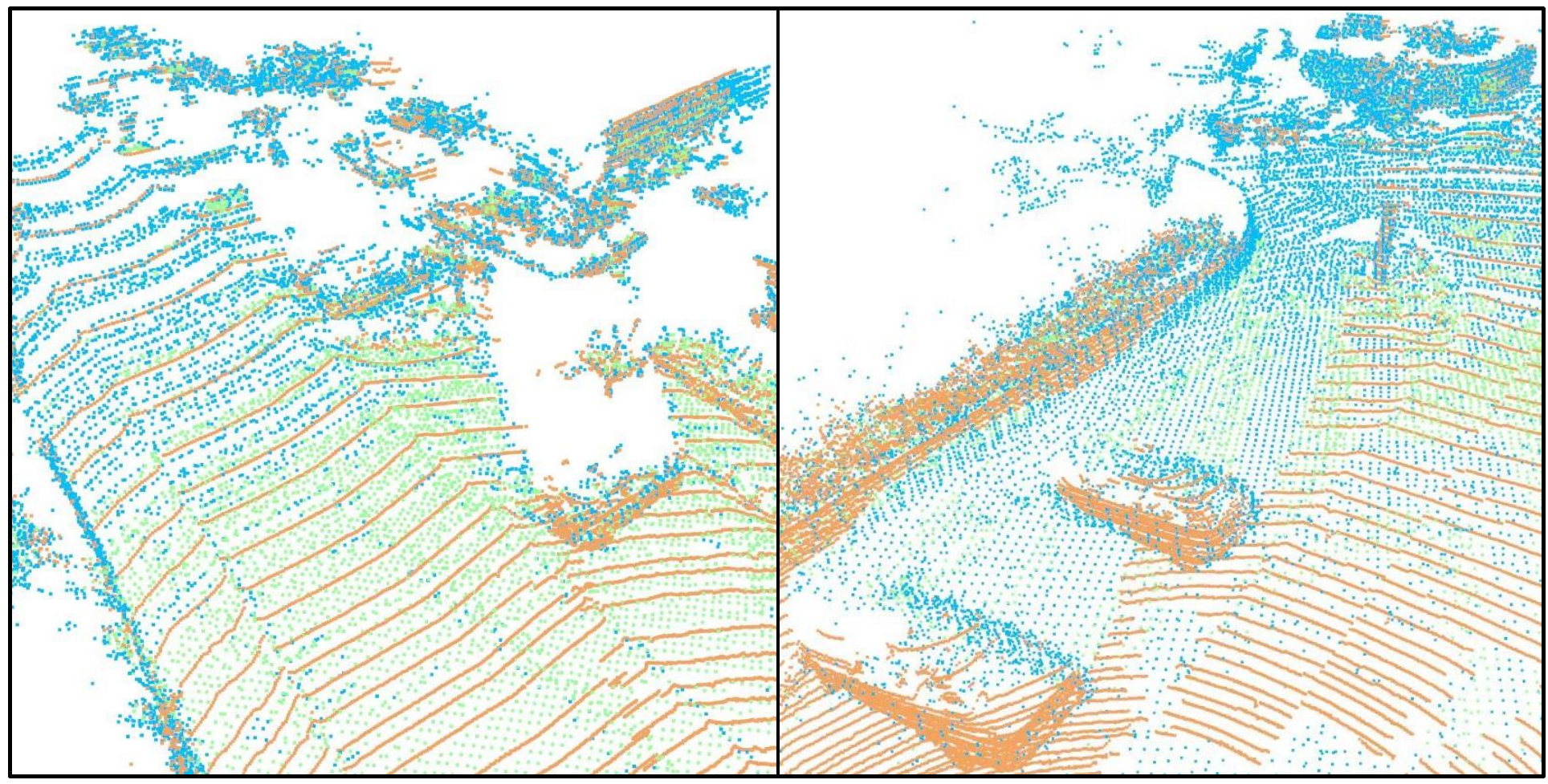}
   \caption{
   The visualization of the attention map of our proposed \textit{Context Activator}, where orange denotes current points, blue denotes activated \textit{Historical Context}, and green denotes deactivated \textit{Historical Context}.
   }
   \label{fig:CAVisual}
\end{figure}

\begin{table}
\small
  \centering
  \scalebox{0.9}{
  \begin{tabular}{c|ccccc}
    \hline
    Distance ($m$) & [0,12) & [12,24) & [24,36) & [36,$\infty$) & all
    \\
    \hline \hline
    Backbone+ & 54.7 & 50.9 & 49.3 & 36.1 & 52.8 \\
    SVQNet (Ours) & 56.5 & 54.5 & 53.5 & 42.3 & 55.3 \\
    $\Delta$ & +1.8 & +3.6 & +4.2 & +6.2 & +2.5
    \\
    \hline
  \end{tabular}}
  \caption{
  The distance-based mIoU results on SemanticKITTI validation set (seq 08, multi-scan phase).
  The mIoU(\%) changes as the distance interval varies from [0, 12) to [36, $\infty$) (unit: meter).
  "all" shows the mIoU results of all points from 0 $m$ to $\infty$.
  Note that only current points are considered in the evaluation.
  }
  \label{tab:extra1}
\end{table}

\subsection{Results on nuScenes}
\label{4.3}

To further test our proposed \textit{SVQNet}, we run experiments on the nuScenes dataset.
As a result, our method ranks $2^{nd}$ on the leaderboard of the semantic segmentation task, in terms of mIoU.
As demonstrated in Tab.~\ref{nuScenes}, compared to JS3C-Net \cite{Yan_Gao_Li_Zhang_Li_Huang_Cui_2021}, Cylinder3D \cite{zhu2021cylindrical} and SPVNAS \cite{10.1007/978-3-030-58604-1_41}, which employed directly stacking strategy according to their implementation, our method receives a performance gain of $7.6\%$, $4\%$ and $3.8\%$, respectively, in terms of mIoU, which shows the superior performance of our proposed \textit{SVQNet} on the semantic segmentation task.

\begin{table*}[!t]
    \centering
    \resizebox{\textwidth}{!}{
\setlength{\tabcolsep}{1.0mm}{      \def\arraystretch{1.2}
    \begin{tabular}{@{}l | c c c c c c c c c c c c c c c c c c c c c c c c c | c c c @{}}
    \hline
    Methods & \rotatebox{90}{road} & \rotatebox{90}{sidewalk} & \rotatebox{90}{parking} & \rotatebox{90}{other ground} & \rotatebox{90}{building} & \rotatebox{90}{car} & \rotatebox{90}{truck} & \rotatebox{90}{bicycle} & \rotatebox{90}{motorcycle} & \rotatebox{90}{other vehicle \,} & \rotatebox{90}{vegetation} & \rotatebox{90}{trunk} & \rotatebox{90}{terrain} & \rotatebox{90}{person} & \rotatebox{90}{bicyclist} & \rotatebox{90}{motorcyclist} & \rotatebox{90}{fence} & \rotatebox{90}{pole} & \rotatebox{90}{traffic sign} & \rotatebox{90}{mov. car} & \rotatebox{90}{mov. bicyclist} & \rotatebox{90}{mov. person} & \rotatebox{90}{mov. motorcyc.} & \rotatebox{90}{mov. truck} & \rotatebox{90}{mov. other veh.} & \rotatebox{90}{mIoU ($\%$)} & \rotatebox{90}{Latency(ms)} & \rotatebox{90}{GPU Memory(MB)}
    \\
    \hline \hline
    Backbone+ & 94.1 & 82.0 & 56.8 & 1.9 & 90.8 & 95.4 & 66.7 & 40.0 & 63.3 & 71.1 & 88.7 & 69.1 & 79.9 & 28.5 & 0.0 & 0.0 & 60.1 & 63.0 & 53.0 & 63.4 & 84.2 & 59.8 & 0.0 & 0.0 & 0.2 & 52.8 & 125 & 8893
    \\
    \hline
    SVQNet(Ours) & 94.9 & 83.7 & 57.0 & 0.3 & 88.6 & 97.3 & 90.0 & 47.1 & 75.8 & 78.4 & 89.4 & 66.9 & 79.1 & 32.0 & 0.0 & 0.0 & 47.8 & 65.5 & 55.5 & 74.7 & 92.5 & 67.3 & 0.0 & 0.0 & 0.0 & 55.3 & 97 & 6854
    \\
    \hline
    $\Delta$ & +0.8 & +1.7 & +0.2 & -1.6 & -2.2 & +1.9 & +23.3 & +7.1 & +12.5 & +7.3 & +0.7 & -2.2 & -0.8 & +3.5 & 0.0 & 0.0 & -12.3 & +2.5 & +2.5 & +11.3 & +8.3 & +7.5 & 0.0 & 0.0 & -0.2 & +2.5 & -28 & -2039
    \\
    \hline
    \end{tabular}}}
    \caption{The per-class results on SemanticKITTI validation set (seq 08, multi-scan phase).
    (mov. denotes moving)}
    \label{tab:extra2}
\end{table*}

\subsection{Ablation Studies}
\label{4.4}

In the ablation study, we conduct several experiments on the validation set (sequence $08$, multi-scan phase) of the SemanticKITTI dataset.

\noindent \textbf{Ablation study on our proposed modules.}
As shown in \cref{tab:ablation}, getting started from our naive multi-scan backbone that stacks all points as input along with $2$ historical frames, we get the baseline performance of $52.8\%$ mIoU.
By enabling the \textit{CA} module, the network gains the most improvement of $1.1\%$ mIoU, which proves that sequence data does have a lot of redundancy and the proposed \textit{CA} module can activate the truly instructive information.
Furthermore, if we activate the \textit{SVAQ} module to enhance features of the current frame, the performance continues to grow $0.9\%$ mIoU, demonstrating that historical knowledge is also crucial to enhance current voxel features.
In the end, \textit{TFI} method is applied to inherit previous features,
and we get a gain of $0.5\%$, which proves the effectiveness to reuse previous computed features with proposed \textit{TFI}.

\noindent \textbf{Ablation study on Context Activator.}
To examine the effectiveness of the proposed \textit{CA}, 
we control the number of activated contextual voxels $M'$ by adjusting the threshold value $S_{th}$ when applying selection on \textit{Historical Context}.
With the growth of $M'$, the provided information is more but the inference time is larger.
As illustrated in \cref{tab:ablation3}, we vary the number from $30k$ to $50k$, finding that \textit{SVQNet} achieves the highest mIoU $55.3\%$ at $M'=40k$.
As $M'$ increases to $45k$, the performance falls, further proving the importance of historical feature selection.
In our experiments mentioned above, we fix $M'=40k$.

\noindent \textbf{Visualization of attention map.}
Additionally, we visualize the attention map to directly understand what our proposed \textit{Context Activator} has learned.
As shown in Fig.~\ref{fig:CAVisual}, we color the current points orange, activated \textit{Historical Context} blue, and deactivated \textit{Historical Context} green.
In this visualization, we fix the number of activated \textit{Historical Context} to about $40k$ by altering the threshold $S_{th}$ mentioned in our method.
Based on the observation, we conclude that, 
1) our proposed \textit{Context Activator} tends to complement the current point clouds at the distant place, where the objects are too sparse to be recognized;
2) our proposed \textit{Context Activator} tends to complete the intricate objects such as cars since the instructive shape priors are significant to semantic segmentation as JS3C-Net \cite{Yan_Gao_Li_Zhang_Li_Huang_Cui_2021} proved.

\noindent \textbf{Comparisons with Backbone+.}
To reveal the source of mIoU gain under the same data input, we conduct experiments and make comparisons with the Backbone+, which directly stacks two historical frames with the current frame as the input of the backbone network.
We show distance-based mIoU results in Tab.~\ref{tab:extra1}.
As the distance interval varies from [0, 12) to [36, $\infty$) (unit: meter), the mIoU results of Backbone+ and ours descend because the farther away the LiDAR scene gets, the more sparse the point clouds become.
However, ours drops less than Backbone+ when distance increases.
And we show the mIoU gap in $\Delta$, revealing our robustness withstanding the distance sparsity of LiDAR.
In addition, Tab.~\ref{tab:extra2} shows the per-class results of Backbone+ and ours, and also the gap.
We can see that, under the input of the same data, ours achieves distinct improvement in intricate objects such as static/moving cars, static/moving people, static trucks, bicycles, motorcycles, and traffic signs.
That reveals the source of our mIoU gain from different categories and the strong ability to distinguish static and moving objects.
Last but not least, we compare ours with Backbone+ regarding latency and GPU memory.
As Tab.~\ref{tab:extra2} demonstrates, under the input of the same data and the same experiment settings, our method is 22.4\% faster than Backbone+, and we also use 22.9\% less GPU memory, benefiting from the proposed information shunt module \textit{STIS} and context selection module \textit{CA} that avoid wasting computation on redundant information.
That exhibits our device-friendly character and the potential for real-time application.

\noindent \textbf{Plugin properties.} 
To prove the plugin properties mentioned in \cref{sec:method}, we perform experiments using SPVCNN~\cite{10.1007/978-3-030-58604-1_41} as our backbone.
The network architecture is demonstrated in the top row of \cref{fig3}.
Under the same input, which is two historical frames along with the current frame, equipping with our modules outperforms the SPVCNN baseline employing stacking strategy by 2.32\% mIoU, as shown in Tab.~\ref{tab:spvcnn}.

\begin{table}
    \small
    \centering
    \begin{tabular}{c|ccc}
    \hline
    SPVCNN & stacking  & + CA\&SVAQ\&TFI \\
    \hline \hline
    mIoU ($\%$)  & 50.04 & \textbf{52.36}\\
    \hline
    \end{tabular}
    \caption{
    The mIoU of SPVCNN (w/o or w/ our modules) on the SemanticKITTI dataset (validation set, multi-scan settings).
    }
    \vspace{-10 pt}
    \label{tab:spvcnn}
\end{table}

\section{Conclusion}
\label{conclu}

We propose \textit{Sparse Voxel-Adjacent Query Network} to concentrate on efficiently extracting 4D spatio-temporal features.
The \textit{Spatio-Temporal Information Shunt} module is proposed to high-efficiently shunt the 4D spatio-temporal information in two groups, \textit{Voxel-Adjacent Neighborhood} and \textit{Historical Context}.
Further, we propose two novel modules, \textit{Sparse Voxel-Adjacent Query} and \textit{Context Activator}, benefiting from the locally \textit{enhancing} information of \textit{Voxel-Adjacent Neighborhood} and globally \textit{completing} information of \textit{Historical Context}.
In addition, \textit{Temporal Feature Inheritance} method is introduced to collect the features preserved in historical frames as an input in the current frame. 
The proposed \textit{SVQNet} reaches state-of-the-art performance on nuScenes and SemanticKITTI leaderboards.

\noindent \textbf{Acknowledgement} 
This research has been made possible by funding support from 1) Deeproute.ai, 2) the Research Grants Council of Hong Kong under the Research Impact Fund project R6003-21, and 3) Natural Science Foundation of China (NSFC) under contract No. 62125106, 61860206003 and 62088102; Ministry of Science and Technology of China under contract No. 2021ZD0109901.


{\small
\bibliographystyle{ieee_fullname}
\bibliography{egbib}
}

\end{document}